\documentclass{article}

\usepackage{PRIMEarxiv}

\usepackage[utf8]{inputenc} 
\usepackage[T1]{fontenc}    
\usepackage{hyperref}       
\usepackage{url}            
\usepackage{booktabs}       
\usepackage{amsfonts}       
\usepackage{nicefrac}       
\usepackage{microtype}      
\usepackage{lipsum}
\usepackage{fancyhdr}       
\usepackage{graphicx}       
\graphicspath{{media/}}     
\usepackage{algorithm}
\usepackage{algpseudocode}
\usepackage{amsmath}

\title{UrbanDIFF: A Denoising Diffusion Model for Spatial Gap Filling of Urban Land Surface Temperature Under Dense Cloud Cover

\author{
  Arya Chavoshi \\
  Maseeh Department of Civil, Architectural, and Environmental Engineering\\
  Jackson School of Geosciences\\
  The University of Texas at Austin \\
  Austin, TX \\
  \texttt{aryachavoshi@utexas.edu} \\
  \And
  Hassan Dashtian \\
  Bureau of Economic Geology \\
  Jackson School of Geosciences\\
  The University of Texas at Austin \\
  Austin, TX \\
  \texttt{hassan.dashtian@beg.utexas.edu} \\
  \And
  Naveen Sudharsan \\
  Department of Earth and Planetary Sciences, 
  Jackson School of Geosciences\\ 
  The University of Texas at Austin \\
  Austin, TX \\
  \texttt{naveens@utexas.edu} \\
  \And
  Dev Niyogi \\
  Maseeh Department of Civil, Architectural, and Environmental Engineering\\
  Department of Earth and Planetary Sciences, 
  Jackson School of Geosciences\\ 
  The University of Texas at Austin \\
  Austin, TX \\
  \texttt{dev.niyogi@utexas.edu} \\
}

}

\begin{document}
\maketitle

\begin{abstract}

Satellite-derived Land Surface Temperature (LST) products are central to surface urban heat island (SUHI) monitoring due to their consistent, near-real-time high resolution coverage over metropolitan regions. However, cloud contamination frequently obscures LST observations, limiting their usability for continuous SUHI analysis. Various reconstruction approaches are being used to provide LST data. Most of the current LST reconstruction approaches are either multi-temporal, relying on cloud-free observations from neighboring times, or multi-sensor, incorporating auxiliary variables from other sensors. These methods require additional information that may be unavailable or unreliable under persistent cloud cover. Purely spatial gap-filling methods provide an alternative, but traditional statistical approaches (e.g., kriging, interpolation) perform poorly when cloud gaps are large or spatially contiguous, while many deep learning–based spatial models produce deterministic reconstructions that degrade sharply under high missingness.

Recent advances in denoising diffusion–based image inpainting models have shown to outperform conventional deep learning and generative adversarial network (GAN)–based approaches, particularly under high levels of missingness and spatially contiguous occlusions, motivating their adoption for spatial LST reconstruction.

In this work, we introduce UrbanDIFF, a purely spatial denoising diffusion model for reconstructing cloud-contaminated urban LST imagery. The model is conditioned on static urban structure-related information, including built-up surface data (city skeleton) and a digital elevation model (DEM). During inference, UrbanDIFF enforces strict consistency with all revealed cloud-free pixels through a supervised refinement step applied before each denoising iteration.

UrbanDIFF is trained and evaluated using NASA MODIS Terra LST data from seven major U.S. metropolitan areas spanning 2002 to 2025. The experiments are conducted on synthetic cloud masks with 20–85\% coverage and varying spatial density, and compared against an interpolation baseline (Telea/OpenCV). UrbanDIFF consistently achieves higher reconstruction fidelity, particularly under dense cloud occlusion, attaining SSIM $\approx 0.89$, RMSE $\approx 1.2 K$, and R² $\approx 0.84$ at 85\% cloud coverage, while exhibiting substantially slower performance degradation than the baseline as cloud mask density increases.  These results demonstrate the potential of denoising diffusion-based models for robust urban LST reconstruction and SUHI analysis. \textbf{Code will be made publicly available upon publication.}

keywords: Land Surface Temperature, Denoising Diffusion Models, Surface Urban Heat Island, Cloud Cover

\end{abstract}

\section{Introduction}

The Land Surface Temperature (LST) derived from the thermal infrared (TIR) satellite is a fundamental variable to quantify and monitor the effect of the Surface Urban Heat Island (SUHI) at regional to global scales due to its relatively high spatiotemporal resolution \cite{Sobrino2013,Lo1997ApplicationOH,Zhou2018,Haashemi2016Seasonal}. However, the operational use of satellite-derived LST products is severely hampered by frequent cloud cover and cloud shadows, resulting in substantial missing LST data \cite{Weng2014ModelingAP}. This limitation is particularly problematic for continuous SUHI monitoring, where spatially and temporally complete LST fields are required.

A broad range of Land Surface Temperature (LST) reconstruction methods have been developed to mitigate cloud-induced data loss. These methods can generally be categorized into four classes: \textit{(1) spatial}, \textit{(2) temporal}, \textit{(3) spatiotemporal}, and \textit{(4) multi-source data fusion--based gap-filling approaches} \cite{mo2021review}.

\textbf{Spatial gap-filling methods} estimate missing pixels using neighboring observed pixels within the same image, often through statistical interpolation techniques and auxiliary static variables such as Digital Elevation Models (DEM) \cite{wu2019reconstructing, zhou2012land}. Although computationally efficient and conceptually simple, these approaches typically perform well only when the cloud gaps are small and spatially interior, and they often fail under heavy or spatially contiguous cloud cover.

\textbf{Temporal gap-filling methods} operate at the pixel level, reconstructing missing values using observations of the same pixel from neighboring time steps via temporal predictive models \cite{crosson2012daily, coops2007estimating, li2014recovering}. Although effective when sufficient cloud-free observations are available, these methods rely on temporal continuity and are limited in cases of persistent cloud cover.

\textbf{Spatiotemporal gap-filling methods} take advantage of spatial and temporal correlations by incorporating information from neighboring pixels across space and time. These approaches have been implemented using statistical models \cite{tan2021reconstruction, pede2018empirical} as well as machine learning--based techniques \cite{Zhao2020Reconstruction}. While generally more accurate than purely spatial methods, spatiotemporal approaches require high-quality LST observations in adjacent time steps, which may not be available under prolonged or widespread cloud conditions.

\textbf{Multi-source data fusion--based methods} reconstruct LST by integrating contemporaneous observations from auxiliary data sources using machine learning or physically informed models. These auxiliary sources commonly include passive microwave measurements that can penetrate clouds \cite{prigent2016toward, catherinot2011evaluation, yoo2020estimation} or atmospheric reanalysis products \cite{fu2019physical, zhu2010enhanced, zhang2021practical}. However, passive microwave data typically suffer from coarse spatial resolution and reduced accuracy at urban scales, while the quality of reanalysis data is highly dependent on the density and distribution of meteorological stations. In regions with sparse station coverage or complex terrain, uncertainties in the reanalysis data can propagate into reconstructed LST fields, thus limiting the precision of the reconstruction.

Due to the strong dependence of multi-source data fusion–based gap-filling methods on the availability and quality of auxiliary variables, there is a clear need for a purely spatial gap-filling model capable of reconstructing LST under heavy cloud occlusion—conditions under which both statistical and conventional deep learning–based spatial methods often fail.

Denoising diffusion–based models have recently emerged as state-of-the-art generative frameworks for image synthesis and inpainting, outperforming variational autoencoders (VAEs) and generative adversarial networks (GANs) in terms of training stability, perceptual quality, and probabilistic formulation \cite{lugmayr2022repaint, zhang2023coherentimageinpaintingusing, corneanu2024latentpaint}. More recently, diffusion-based models have demonstrated strong performance in optical satellite imagery inpainting \cite{zou2024diffcr, panboonyuen2025satdiff,cai2025fusing}, and TIR-derived LST cloud removal \cite{leher2025denoising}, often surpassing conventional inpainting approaches in perceptual fidelity and structural consistency, suggesting their strong potential for satellite-based weather and climate data reconstruction. 

In this study, we introduce \textbf{UrbanDIFF}, a denoising diffusion--based model specifically designed for single-image spatial gap filling of MODIS LST over urban regions. Importantly, UrbanDIFF is formulated as a \textit{purely spatial} reconstruction model: it relies solely on the revealed LST pixels within a single masked image, together with two static auxiliary conditions—the urban morphological skeleton and a digital elevation model (DEM). This design enables us to isolate and evaluate the intrinsic generative capability of diffusion models for LST reconstruction under heavy cloud \emph{occlusion}, independent of temporal priors or cross-sensor information. Accordingly, UrbanDIFF is intended as a methodological investigation of denoising diffusion--based spatial gap-filling performance, rather than a complete operational solution for producing temporally continuous LST products.

The objectives of this study are the following.
\begin{enumerate}
    \item Develop a reliable and computationally efficient denoising diffusion model for purely spatial gap filling of urban LST, with a particular focus on heavy cloud cover scenarios.
    \item Evaluate the ability of the proposed model to reconstruct spatially coherent LST fields under a wide range of cloud coverage levels and occlusion patterns, using statistical reconstruction metrics and surface urban heat island (SUHI) estimation error.
    \item UrbanDIFF benchmark against a widely used statistical interpolation baseline, with an emphasis on comparative performance under high cloud coverage conditions.
    \item Investigate the influence of key denoising diffusion hyperparameters and identify optimal model configurations across different levels of cloud contamination.
    \item Provide qualitative assessments of UrbanDIFF reconstructions to examine spatial structure recovery under severe cloud occlusion.
\end{enumerate}

This work provides a systematic evaluation of denoising diffusion-based generative models for single-image LST gap filling in urban environments, and offers insights into their potential for advancing satellite-based SUHI monitoring.

\section{Methods and Data}
\begin{figure}
    \centering
    \includegraphics[width=1\linewidth]{method.png}
    \caption{Methodological overview of \textbf{UrbanDIFF}. 
    \textbf{(a)} Schematic of the overall inference pipeline, illustrating the iterative denoising process combined with pixel-level supervised refinement across diffusion timesteps. At each timestep, denoising steps are interleaved with gradient-based updates that enforce consistency with revealed (cloud-free) pixels. 
    \textbf{(b)} Details of the supervised pixel-guided refinement step, including the loss function defined on revealed pixels and the corresponding gradient update applied before each denoising iteration. 
    \textbf{(C)} Conditional denoising step of the diffusion process, showing the probabilistic formulation of the reverse diffusion transition.}

    \label{fig:1}
\end{figure}
\begin{figure}
    \centering
    \includegraphics[width=1\linewidth]{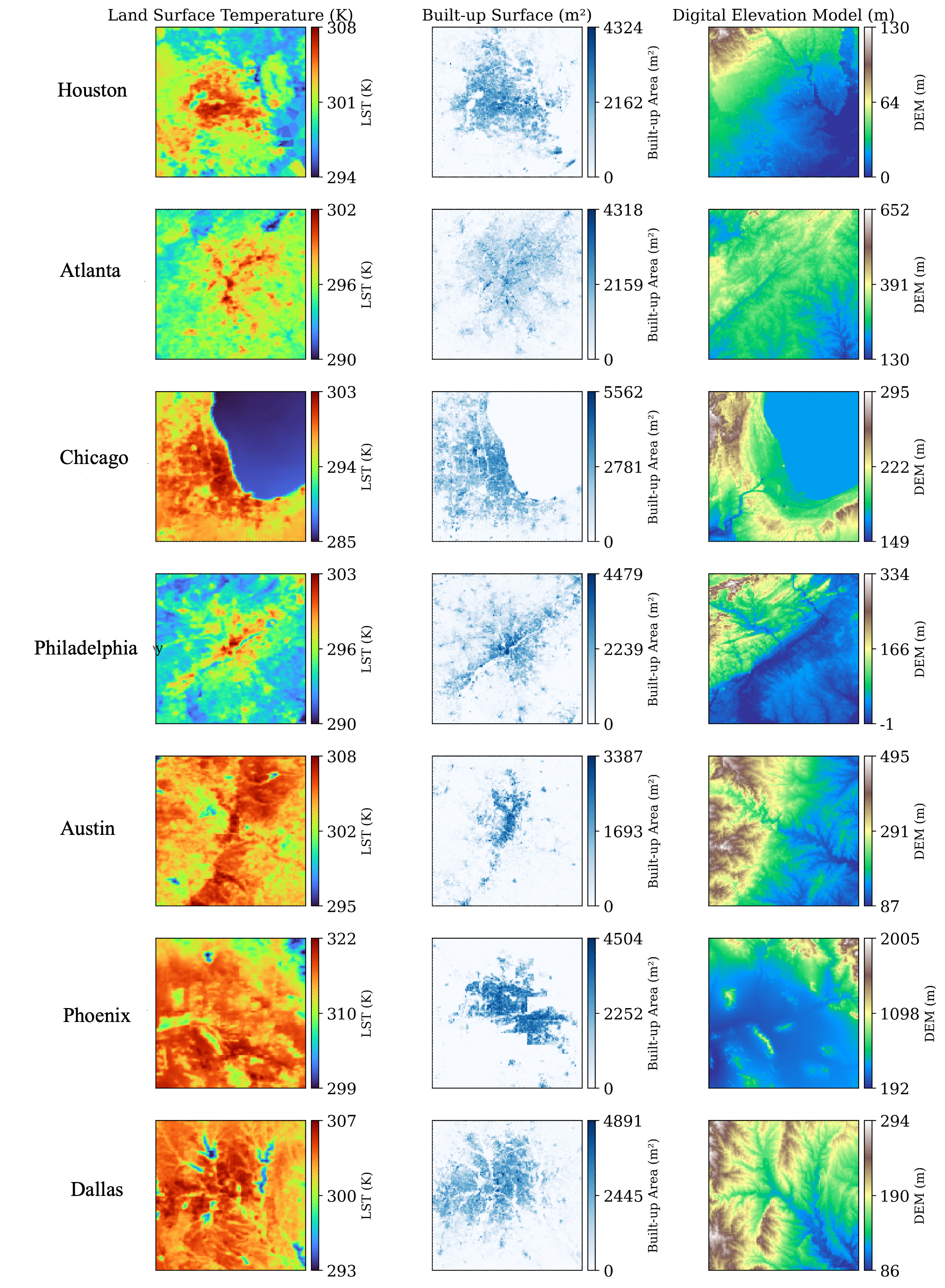}
   \caption{Overview of the datasets used in this study, showing MODIS LST for selected urban regions and the corresponding static conditioning variables (built-up surface and DEM).}
    \label{fig:2}
\end{figure}
\subsection{Training}
Diffusion-based generative models consist of two complementary processes: 
a forward noise-perturbation process and a reverse denoising process. 
During training, the forward diffusion process gradually corrupts clean data samples by adding Gaussian noise over many steps, eventually transforming them into standard Gaussian noise. A neural surrogate model is then trained on these progressively noised samples to predict and separate the noise component from the underlying clean signal.
During inference, the reverse diffusion process uses this learned surrogate model to iteratively remove noise—step by step—thereby inverting the corruption process and reconstructing new samples drawn from the data distribution.

In this work, we adopted the Denoising Diffusion Probabilistic Model (DDPM) \cite{ho2020denoising} as our backbone for the forward noise perturbation process. 
The forward diffusion process adds a small amount of Gaussian noise at each step:

\begin{equation}
    [x_t \mid x_{t-1}] := \mathcal{N}\!\left(\sqrt{1-\beta_t}\, x_{t-1},\, \beta_t I \right),
\end{equation}

where $[a]$ denotes the probability distribution of the random variable $a$, $x_t$ is the corrupted sample at timestep $t$ where $t = 0:T$ ($x_0$: clean image; $x_T$: standard Gaussian noise),
and $\beta_t$ is the diffusion rate drawn from a predefined variance schedule. 
We use a linear variance schedule with 1000 steps ($T = 1000$), where 
$\beta_0 = 10^{-4}$ and $\beta_{1000} = 2\times 10^{-2}$.

Because Gaussian perturbations compose analytically, 
the forward process has a closed-form solution that allows sampling $x_t$ directly 
from the clean sample $x_0$ at any timestep $t$:

\begin{equation}
    x_t = \sqrt{\bar{\alpha}_t}\, x_0 
    + \sqrt{1 - \bar{\alpha}_t}\, \epsilon_t,
\end{equation}

where $\epsilon_t \sim \mathcal{N}(0, I)$, 
$\alpha_t = 1 - \beta_t$, 
and $\bar{\alpha}_t = \prod_{s=1}^{t} \alpha_s$ is the cumulative noise attenuation factor.

The reverse diffusion process defines a denoising Markov chain:

\begin{equation}
    [x_{t-1} \mid x_t, x_0] := \mathcal{N}(\tilde{\mu}_t,\, \tilde{\beta}_t I),
\end{equation}

with posterior mean

\begin{equation}
    \tilde{\mu}_t
    =
    \frac{1}{\sqrt{\alpha_t}}
    \left(
        x_t 
        - 
        \frac{1-\alpha_t}{\sqrt{1-\bar{\alpha}_t}}\, \epsilon_t
    \right).
\end{equation}

Since the true noise $\epsilon_t$ is unknown during denoising, 
$\epsilon_\theta(x_t, C, t)$ is trained as a neural surrogate model to estimate it, 
given the noisy sample $x_t$, optional conditioning variables $C$, 
and the timestep $t$. 
The training objective is the standard DDPM denoising loss defined as:

\begin{equation}
    l = \lVert \epsilon_t - \epsilon_{\theta}(x_t, C, t) \rVert^2.
\end{equation}

We adopt $\epsilon$-prediction parameterization for the denoiser, consistent with DDPM and the DPMSolver++ scheduler.

\subsection{Inference}
\label{subsec:inference}

The objective of the inference process is to generate a cloud–free reconstruction of an LST image given its partially observed version $x_0 \odot M$, where $M$ is a known binary cloud mask ($M = 1$ in revealed pixels and $M = 0$ in cloud gaps), inferred from MODIS QA flags and applied synthetically during evaluation. The diffusion model is also conditioned on external variables $C$, so the inference objective is $[x_0 \,|\, x_0 \odot M, C]$.

Using the trained denoiser $\epsilon_{\theta}(x_t, C, t)$, we adapt components from both COPAINT--TT \cite{zhang2023coherentimageinpaintingusing} and RePaint \cite{lugmayr2022repaint} to perform guided inpainting during the reverse diffusion process. The key idea is to (i) refine the sample at selected timesteps so that its one–step clean prediction matches the revealed pixels, and (ii) project the refined sample back onto the known pixels, similar to RePaint.

\paragraph{Supervised Gradient Refinement (COPAINT--TT Style).}
At selected denoising timesteps $t = T, \dots, 1$, we insert a fixed number of supervised gradient updates before the denoising update. For each timestep $t$, we perform $G$ refinement steps:
\begin{equation}
    x_t^{g} = x_t^{g-1} - \gamma \, \nabla l_t^{g-1},
    \label{eq:grad_step}
\end{equation}
where $x_t^{g}$ denotes the intermediate refined sample after $g$ gradient steps and $\gamma$ is the step size.

The supervised loss penalizes deviations from the ground truth on revealed pixels:
\begin{equation}
    l_t^{g-1}
    =
    \left\| 
        \bar{x}_{0,t}^{g-1} \odot M \;-\; x_0 \odot M 
    \right\|^2,
    \label{eq:loss}
\end{equation}
where $\odot$ denotes elementwise multiplication, and
\begin{equation}
    \bar{x}_{0,t}^{g-1}
    =
    \frac{
        x_t^{g-1}
        - \sqrt{1 - \bar{\alpha}_t}\,
          \epsilon_{\theta}(x_t^{g-1}, C, t)
    }{
        \sqrt{\bar{\alpha}_t}
    }
    \label{eq:x0_bar}
\end{equation}
is the one–step clean-sample estimate at timestep $t$.

\paragraph{RePaint-Style Projection.}
After the gradient refinement, we enforce consistency with the known pixels. We first generate a noise-perturbed partially observed sample:
\[
    (x_0 \odot M)_t 
    := 
    \sqrt{\bar{\alpha}_t}\,(x_0 \odot M)
    + 
    \sqrt{1 - \bar{\alpha}_t}\,\epsilon_t .
\]
We then overwrite the refined sample $x_t^{G}$ on the revealed pixels:
\[
    \hat{x}_t^{G} 
    =
    (x_0 \odot M)_t \odot M
    \;+\;
    x_t^{G} \odot (1 - M).
\]

\paragraph{Denoising Step.}
Conditioned on $\hat{x}_t^{G}$, the reverse diffusion transition is given by
\begin{equation}
    [x_{t-1}^{0} \mid \hat{x}_t^{G}]
    \sim
    \mathcal{N}\!\left(
        \mu_{\theta}(\hat{x}_t^{G}, C, t),\;
        \tilde{\beta}_t I
    \right),
    \label{eq:reverse_transition}
\end{equation}
with mean
\begin{equation}
    \mu_{\theta}(\hat{x}_t^{G}, C, t)
    =
    \frac{1}{\sqrt{\alpha_t}}
    \left(
        x_t^{G}
        -
        \frac{1-\alpha_t}{\sqrt{1 - \bar{\alpha}_t}}
        \epsilon_{\theta}(\hat{x}_t^{G}, C, t)
    \right).
    \label{eq:reverse_mean}
\end{equation}

\paragraph{Inference Efficiency.}
To reduce computation, no refinement is applied in every timestep. Instead, we introduce a guidance stride $\tau$: supervised updates are performed only when $t \bmod \tau = 0$.

\begin{algorithm}[H]
\caption{UrbanDIFF Inference with Supervised Gradient Refinement}
\label{alg:urbandiff}
\begin{algorithmic}[1]

\State \textbf{Input:} Trained denoiser $\epsilon_{\theta}$, conditions $C$, partially observed image $x_0 \odot M$, mask $M$, total steps $T$, refinement steps $G$, stride $\tau$, step size $\gamma$
\State \textbf{Output:} Reconstructed sample $\hat{x}_0$

\State Initialize $x_T^{0} \sim \mathcal{N}(0, I)$

\For{$t = T, T-1, \dots, 1$}

    \If{$t \bmod \tau = 0$}
        \State \textbf{// (A) Supervised Gradient Refinement}
        \For{$g = 1$ to $G$}
            \State Compute clean prediction using \eqref{eq:x0_bar}
            \State Compute masked loss using \eqref{eq:loss}
            \State Update sample using \eqref{eq:grad_step}
        \EndFor
        \State $x_t^{G} \gets x_t^{G}$ 
    \Else
        \State $x_t^{G} \gets x_t^{0}$ 
    \EndIf

    \State \textbf{// (B) RePaint-style Projection}
    \State Generate perturbed observed sample $(x_0 \odot M)_t$
    \State $\hat{x}_t^{G} = (x_0 \odot M)_t \odot M + x_t^{G} \odot (1 - M)$

    \State \textbf{// (C) Reverse Diffusion Step}
    \State Sample $x_{t-1}^{0}$ using \eqref{eq:reverse_transition}--\eqref{eq:reverse_mean}

\EndFor

\State \Return $x_0^{0}$ (or the refined version if refinement occurs at $t=1$)

\end{algorithmic}
\end{algorithm}

\subsection{Data Collection and Preprocessing}

We used the \textbf{NASA MODIS MOD11A1.061 Terra Land Surface Temperature (LST) and Emissivity Daily Global 1\,km} product \cite{wan2015myd11a1} to construct a long-term dataset of urban thermal conditions across seven major U.S.\ metropolitan regions: \textbf{Austin, Houston, Dallas, Atlanta, Phoenix, Chicago, and Philadelphia}. These cities were selected to span a wide range of \emph{climates} (humid subtropical, semi-arid, desert, continental) and \emph{geographical settings} (coastal and inland), allowing generalization across diverse urban morphologies.

Daily Terra LST scenes from \textbf{2002--2025} were retrieved for each city. Cloud-free or near--cloud-free scenes were identified using the accompanying MODIS quality assurance (QA) flags. For every valid scene, we extracted a \textbf{128 $\times$ 128 pixel} window (\(\sim 128 \,\text{km} \times 128 \,\text{km}\)) centered on the urban core to maintain a consistent spatial framing. After QA filtering and cropping, we obtained approximately \textbf{12{,}500 images} across all seven regions. These were randomly divided into \textbf{80\% for training} and \textbf{20\% for testing}, ensuring temporal and spatial diversity in both splits.

\subsubsection{Conditioning Variables}

To provide structural and topographic context for the generative diffusion model, we incorporated two static geospatial layers as conditioning variables \(C\):

\begin{itemize}
    \item \textbf{Built-up Surface:} We used the \textbf{GHSL Built-up Surface 1975--2030 (Release P2023A)} \cite{florio2025ghs} and selected the year--2020 layer as the representative built environment for each city. The original 30\,m resolution product was aggregated to the MODIS 1\,km grid via mean resampling to achieve spatial alignment with the LST data.
    
    \item \textbf{Digital Elevation Model:} We used the \textbf{Copernicus DEM GLO-30} (30\,m) \cite{CopernicusDEMGLO30_2023}. This dataset was upscaled to 1\,km resolution using bilinear resampling, providing elevation information that influences terrain-driven thermal variability.
\end{itemize}

The final conditioning tensor; therefore, includes (i) the 1\,km built-up intensity and (ii) the 1\,km elevation, both co-registered to the LST grid. These layers supply the model with the structural layout and topographic context necessary to learn spatially coherent urban heat patterns. Figure~\ref{fig:2} illustrates the average MODIS LST maps for the selected urban regions, together with the corresponding static conditioning variables.

\subsection{Evaluation, Hyperparameter Optimization, and Comparison}
\subsubsection{\textbf{Synthetic Cloud Generation for Test Data}}
To evaluate UrbanDIFF's spatial gap-filling performance, we applied \emph{synthetic cloud masks} to the held-out test images. The cloud masks were generated by randomly sampling the seed blobs and iteratively expanding them to create realistic, irregular cloud structures. The synthetic clouds are characterized by three key attributes: (i) \textbf{cloud coverage}, which controls the fraction of pixels obscured; (ii) \textbf{octaves}, which represent the spatial frequency and density of the cloud patterns, where higher octaves create larger contiguous gaps that are more challenging to reconstruct; and (iii) \textbf{wind direction}, representing the directions of the clouds. 
\subsubsection{\textbf{Hyperparameter Optimization Setup}}
To determine the optimal UrbanDIFF hyperparameters for each cloud condition (defined by cloud coverage (cc) and number of octaves), we conducted a comprehensive hyperparameter sweep. For cloud coverage levels $\{0.2, 0.5, 0.85\}$, and octaves $\{2, 6, 10\}$, we evaluated diffusion timesteps $T \in \{20, 40, 70, 100, 150, 180\}$ and guidance strides $\tau \in \{1, 2, 3, 4\}$. This resulted in a total of 216 hyperparameter configurations.

We fixed the supervised refinement parameters to $G = 1$ and $\gamma = 10$, as increasing the number of gradient steps or modifying the step size often leads to computational instability during inference.

All hyperparameter experiments were performed on the Texas Advanced Computing Center (TACC) Lonestar6 system, using 8 compute nodes equipped with 3 $\times$ A100 GPUs each.

\subsubsection{Baseline Interpolation Model and Evaluation Setup.}
To benchmark UrbanDIFF, we compared its performance against a baseline interpolation algorithm \cite{telea2004image} implemented using the \texttt{OpenCV} library \cite{bradski2000opencv}. This baseline provides a conventional spatial interpolation approach against which the benefits of diffusion-based gap filling can be quantified.

For evaluation, we generated a comprehensive set of \textbf{100 synthetic cloud types} combining multiple factors that influence the difficulty of filling the gap: cc levels \((0.2,\, 0.4,\, 0.5,\, 0.7,\, 0.85)\), cloud octaves \((2,\, 4,\, 6,\, 8,\, 10)\), and wind directions \((0^{\circ},\, 90^{\circ},\, 135^{\circ},\, 180^{\circ})\). Each unique combination of these parameters corresponds to a distinct cloud mask.

We applied every cloud type to the held-out test dataset and evaluated both UrbanDIFF (using the optimized hyperparameter configuration for the selected cloud level) and the baseline interpolation model under identical masking conditions. This exhaustive setup allowed us to quantify model robustness across a wide spectrum of cloud densities, structural complexities, and spatial orientations.

The evaluation process was executed on TACC Lonestar6 system, using 12 compute nodes equipped with 3 $\times$ A100 GPUs each.

\subsubsection{\textbf{Metrics}}

We evaluate UrbanDIFF and the baseline interpolation model using five commonly used image quality and regression metrics: \textbf{SSIM}, \textbf{RMSE}, \textbf{PSNR}, \textbf{SUHI Error}, and \textbf{R\textsuperscript{2}}. Let $x$ denote the ground-truth LST image, $\hat{x}$ the reconstructed (gap-filled) image. All comparisons are computed only over the masked regions:
\[
x_m = x \odot (1-M), \qquad \hat{x}_m = \hat{x} \odot (1-M),
\]. 

Let $N_m$ be the number of masked pixels: $N_m = \sum (1-M)$.

\paragraph{Structural Similarity Index Measure (SSIM).}
SSIM measures perceptual similarity using masked means, variances, and covariances:
\begin{equation}
\mathrm{SSIM}(x_m, \hat{x}_m) 
= \frac{(2\mu_{x_m}\mu_{\hat{x}_m} + C_1)\left(2\sigma_{x_m\hat{x}_m} + C_2\right)}
{(\mu_{x_m}^2 + \mu_{\hat{x}_m}^2 + C_1)\left(\sigma_{x_m}^2 + \sigma_{\hat{x}_m}^2 + C_2\right)},
\end{equation}
where $\mu_{x_m}, \mu_{\hat{x}_m}$ are means over masked pixels,  
$\sigma_{x_m}^2, \sigma_{\hat{x}_m}^2$ are masked variances,  
and $\sigma_{x_m\hat{x}_m}$ is their masked covariance.

\paragraph{Root Mean Squared Error (RMSE).}
RMSE is computed only on masked pixels:
\begin{equation}
\mathrm{RMSE} = 
\sqrt{\frac{1}{N_m} \sum_{i \in M} (x_i - \hat{x}_i)^2 }.
\end{equation}

\paragraph{Peak Signal-to-Noise Ratio (PSNR).}
PSNR is defined from the masked MSE:
\begin{equation}
\mathrm{PSNR} = 
10 \log_{10}\left( \frac{\mathrm{MAX}^2}{\mathrm{MSE}_m} \right),
\end{equation}
where  
\[
\mathrm{MSE}_m = \frac{1}{N_m}\sum_{i \in M}(x_i - \hat{x}_i)^2,
\]
and $\mathrm{MAX}$ is the maximum possible LST value (from cloud-free MODIS range).

\paragraph{Coefficient of Determination (R\textsuperscript{2}).}
$R^2$ is also computed only over masked pixels:
\begin{equation}
R^2 = 1 - 
\frac{\sum_{i \in M} (x_i - \hat{x}_i)^2}
     {\sum_{i \in M} (x_i - \bar{x}_m)^2},
\end{equation}
where $\bar{x}_m$ is the mean of the ground-truth values over masked pixels.

\paragraph{Surface Urban Heat Island Estimation Error (SUHI Error).}
SUHI is computed from the reconstructed image (SUHI is a property of the full scene).  
Let $\mathcal{U}$ denote urban pixels and $\mathcal{R}$ rural pixels:
\[
\mathrm{SUHI}(x) = 
\frac{1}{|\mathcal{U}|}\sum_{i \in \mathcal{U}} x_i 
- 
\frac{1}{|\mathcal{R}|}\sum_{j \in \mathcal{R}} x_j.
\]
SUHI Error compares the reconstructed and true SUHI intensities:
\begin{equation}
\mathrm{SUHI\ Error} = 
|\frac{ \mathrm{SUHI}(x) - \mathrm{SUHI}(\hat{x})}{\mathrm{SUHI}(x)}|\times100
\end{equation}

For hyperparameter optimization, we compared different configurations using a composite normalized score (NS), defined as
\begin{equation}
    NS 
    = 
    0.18\, NS_{\mathrm{SSIM}}
    + 0.18\, NS_{\mathrm{RMSE}}
    + 0.18\, NS_{\mathrm{PSNR}}
    + 0.18\, NS_{R^2}
    + 0.18\, NS_{\mathrm{SUHI}}
    + 0.10\, NS_{\mathrm{inference\;time}}.
\end{equation}

Each component $NS_{\mathrm{metric}}$ is computed by normalizing the metric values across all tested configurations:
\begin{equation}
    NS_{\mathrm{metric}}
    =
    \frac{\mathrm{metric} - \min(\mathrm{metric})}
         {\max(\mathrm{metric}) - \min(\mathrm{metric})}.
\end{equation}
For metrics where lower values indicate better performance (RMSE, SUHI error, inference time), the normalized score is inverted as $1-NS_{metric}$ before aggregation.

\subsection{Inference and Training Configurations}

The denoiser network $\epsilon_{\theta}$ is a U-Net developed by \cite{DBLP:journals/corr/RonnebergerFB15}. The model operates at spatial resolution $128 \times 128$ and takes three input channels. The U-Net consists of \textbf{four resolution levels}, with channel widths of \textbf{64, 128, 256, and 256} in the downsampling path. Each level contains \textbf{two convolutional layers} per block. Self-attention is incorporated at the two deepest levels of the U-Net (the 256-channel blocks), allowing the model to capture long-range spatial dependencies important in urban thermal structure. The upsampling path mirrors the downsampling path and also includes attention at the corresponding high-level feature blocks. All attention layers use \textbf{8 attention heads} with a dropout rate of \textbf{0.1}. The model outputs a single-channel residual prediction corresponding to the noise estimate in the diffusion process.

Training of $\epsilon_{\theta}$ was performed for \textbf{75 epochs} using a batch size of \textbf{16} and an initial learning rate of $10^{-4}$, decayed by a factor of \textbf{0.9} every two epochs.

\paragraph{Inference Settings.}
During inference, the number of denoising timesteps and guidance stride were chosen according to the hyperparameter optimization test. We applied one \textbf{supervised gradient step} per update ($G=1$) with a step size of \textbf{$\gamma = 10$}.

\paragraph{Denoising Scheduler.}
For the reverse diffusion process, we used the \textbf{DPMSolver++} scheduler \cite{lu2022dpmsolverfastodesolver} with third-order solver updates, epsilon prediction type, Karras noise sigmas, and trailing timestep spacing. This high-order solver provides improved stability and reconstruction quality at moderate $T$. Both the U-Net model and the DPMSolver++ scheduler are implemented from the HuggingFace Diffusers library \cite{von-platen-etal-2022-diffusers}.

\section{Results and Discussion}
\subsection{Hyperparameter Selection}
\begin{figure}
    \centering
    \includegraphics[width=0.9\linewidth]{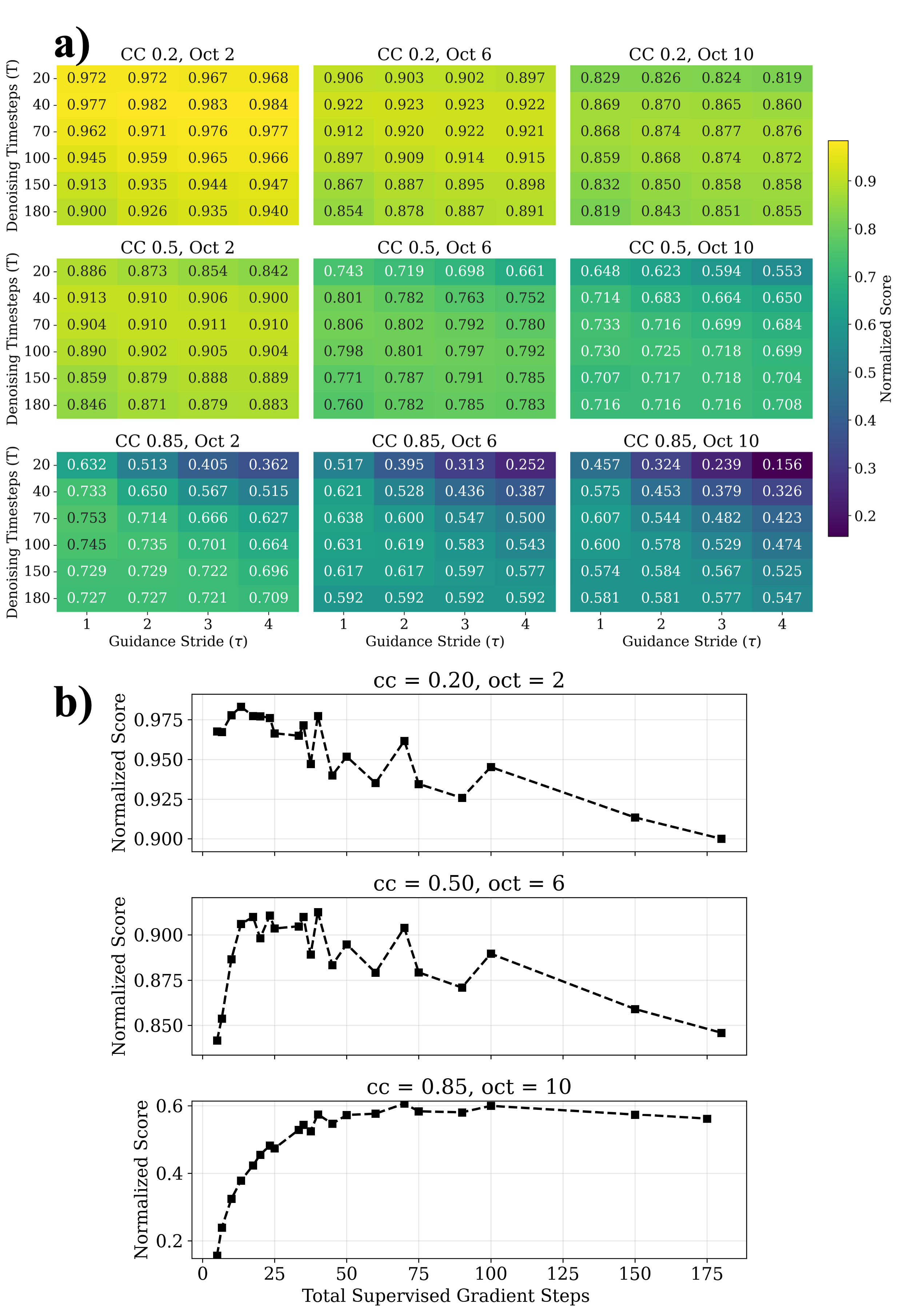}
    \caption{Effect of hyperparameter variations on UrbanDIFF's normalized performance score. The normalized score is computed as a weighted average of normalized SSIM, RMSE, $R^2$, SUHI error, and per-image inference time. 
    \textbf{(a)} Heatmap of the normalized score across synthetic cloud conditions, defined by cloud coverage (cc) and cloud octaves, as a function of the number of denoising timesteps ($T$) and the guidance stride ($\tau$). 
    \textbf{(b)} Influence of the total number of supervised gradient steps ($T/\tau$) on the normalized score for three representative cloud coverage levels.}

    \label{fig:3}
\end{figure}

Figure~\ref{fig:3} summarizes the effect of hyperparameter choices on the UrbanDIFF normalized performance score across three cloud‐cover levels (cc = 0.20, 0.50, 0.85) and 3 octave settings (2, 6, 10). For low cloud cover (cc = 0.20), the best‐performing configurations were $(T,\tau) = (40,4)$ for octave~2, $(40,3)$ for octave~6, and $(70,3)$ for octave~10. As seen in the heatmaps and the trend plot (Figure~\ref{fig:3}b), for cc = 0.20, increasing the number of denoising timesteps or using smaller guidance strides---both of which increase the total number of supervised gradient steps---consistently reduced the performance. Optimal performance was achieved when the total number of supervised gradient steps lay between 15 and~40.

For moderate cloud cover (cc = 0.50), the optimal settings shifted toward $(T,\tau) = (40,1)$ for octaves~2 and $(70,1)$ for octaves~6 and~10. Unlike the cc = 0.20 case, performance no longer decreased monotonically with additional refinement. Instead, performance initially improved, peaked around 40--50 supervised gradient steps, and declined more gradually thereafter (Figure~\ref{fig:3}b, middle panel).

For high cloud cover (cc = 0.85), all octave configurations shared the same optimum: $(T,\tau) = (70,1)$. As illustrated in both the heatmaps and Figure~\ref{fig:3}b (bottom panel), performance increased steadily with more refinement steps until plateauing at around 70 steps. This pattern is the opposite of the low cloud‐cover case, indicating that for highly occluded inputs stronger guidance and more intensive refinement are beneficial.

These findings are consistent with observations from prior diffusion inpainting ablation studies. Lugmayr et al.\ \cite{lugmayr2022repaint} showed that allocating additional computational budget to slowing down the diffusion process (more denoising timesteps) yields little benefit. They further demonstrated that very small step sizes (jump length $j=1$), equivalent to excessive refinement, tend to produce overly smooth or blurry outputs, mirroring our observations that overly strong refinement may harm performance. Similarly, Zhang et al.\ \cite{zhang2023coherentimageinpaintingusing} reported that increasing the number of supervised gradient refinement steps $G$ does not always yield better results in COPAINT-TT method; excessively large $G$ introduces overfitting to revealed pixels and ultimately degrades global reconstruction quality. This aligns with our findings, where too many refinement steps (total gradient steps) reduce performance. 

Collectively, prior diffusion-based inpainting ablation studies corroborate our interpretation that low corruption regimes favor limited refinement and fewer denoising steps, while higher corruption regimes require stronger guidance and extended refinement, with over-refinement degrading performance across all settings.

\subsection{Comparison with Baseline}
\begin{figure}
    \centering
    \includegraphics[width=1\linewidth]{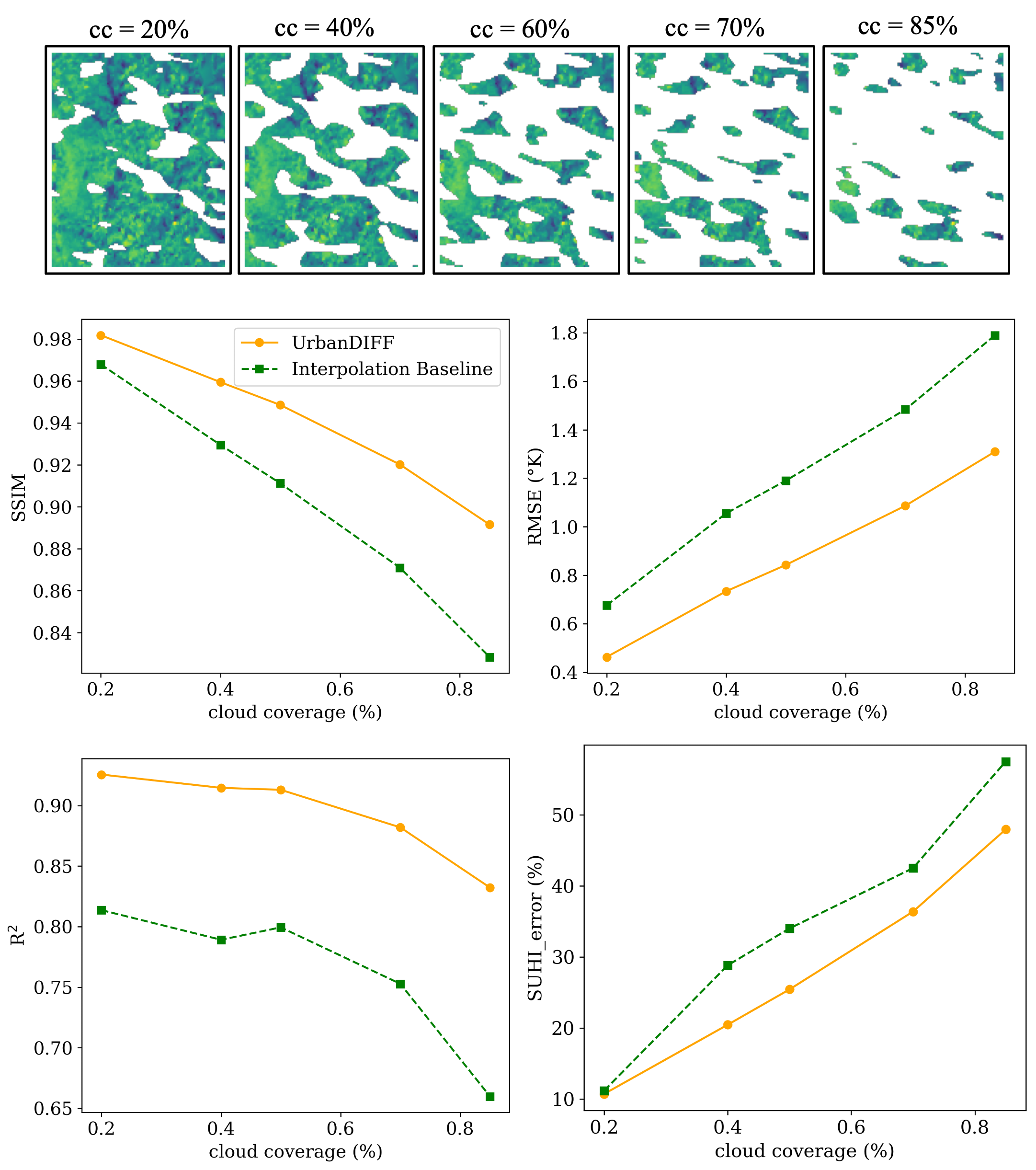}
    \caption{Influence of synthetic cloud coverage on the gap-filling performance of UrbanDIFF and the baseline interpolation model, evaluated using SSIM, RMSE, $R^2$, and SUHI error.}
    \label{fig:4}
\end{figure}

\begin{figure}
    \centering
    \includegraphics[width=1\linewidth]{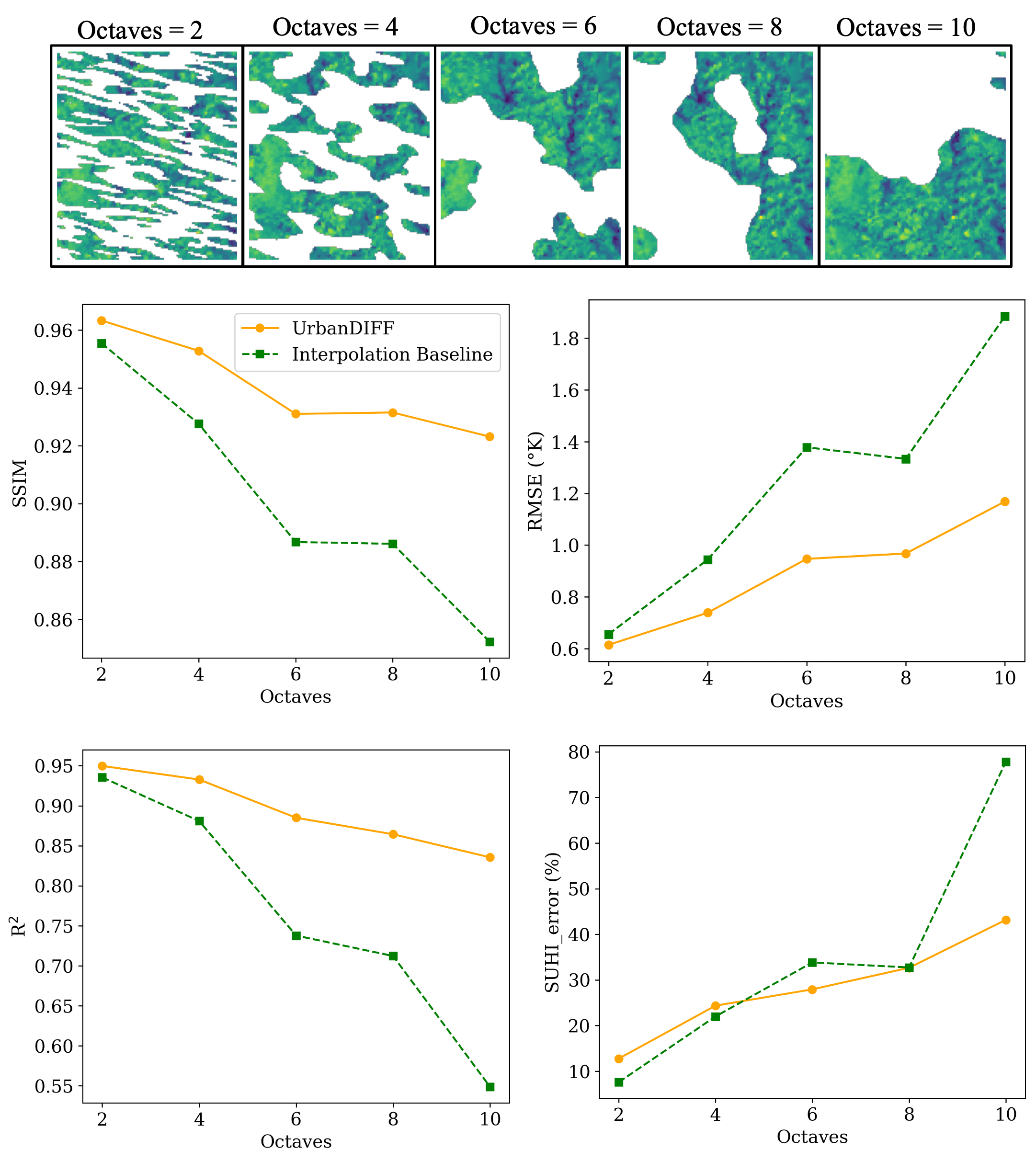}
    \caption{Influence of synthetic cloud mask octaves (mask density) on the gap-filling performance of UrbanDIFF and the baseline interpolation model, evaluated using SSIM, RMSE, $R^2$, and SUHI error.}
    \label{fig:5}
\end{figure}

Figures~\ref{fig:4} and~\ref{fig:5} summarize how the performance of UrbanDIFF and the baseline interpolation model change under increasing reconstruction difficulty, measured through cloud cover and cloud octaves (cloud density). Across all cloud covers from 20\% to 85\%, UrbanDIFF consistently outperforms the baseline, achieving higher SSIM and $R^2$, and lower RMSE and SUHI Error. In addition, the performance gap between the two models widens substantially as cloud contamination becomes more severe.

For SSIM, the difference between UrbanDIFF and the baseline increases from approximately 1\% at 20\% cloud cover (98\% vs.\ 97\%) to about 7\% at 85\% cloud cover (90\% vs.\ 83\%), representing a substantial widening of the performance gap. A similar pattern is observed for $R^2$: the gap increases from 11 percentage points at 20\% cloud cover (92\% vs.\ 81\%) to 18 points at 85\% cloud cover (84\% vs.\ 66\%), corresponding to a 72\% increase in performance separation. RMSE also shows a growing divergence, increasing from \(-0.23\) at 20\% cloud cover (0.41 vs.\ 0.64) to \(-0.55\) at 85\% cloud cover (1.22 vs.\ 1.75), more than doubling the improvement margin. For SUHI\_Error, UrbanDIFF consistently achieves lower error, although the gap remains relatively stable across cloud covers.

A similar pattern emerges when examining cloud octaves, which control the spatial density and contiguity of gaps. At low octaves (octave 2), the SSIM difference is modest (96.5\% vs.\ 95.5\%, a 1\% gap), but at octave 10 it grows to 8\% (93\% vs.\ 85\%), an eight-fold increase. For $R^2$, the models are nearly identical at low octaves (95\% vs.\ 94.5\%), whereas at octave 10 the baseline collapses dramatically (85\% vs.\ 55\%). RMSE similarly diverges from a small difference (0.61 vs.\ 0.66) at octave 2 to a large gap (1.18 vs.\ 1.85) at octave 10. SUHI\_Error remains comparable until octave 8, but at octave 10 UrbanDIFF maintains substantially lower error (40\% vs.\ 80\%).

Overall, these results demonstrate that while UrbanDIFF consistently outperforms the interpolation baseline across all pixel-level metrics, its advantage becomes most pronounced under the most challenging conditions—high cloud coverage and high cloud octaves—where interpolation degrades sharply due to limited contextual information and the difficulty of reconstructing large, contiguous gaps. UrbanDIFF’s ability to recover coherent spatial structure under these severe occlusions highlights the strengths of diffusion-based modeling relative to traditional interpolation approaches.
More broadly, prior studies indicate that both statistical and deep learning–based image inpainting methods exhibit substantial performance degradation as mask coverage and spatial contiguity increase. For example, Suvorov et al.\ \cite{suvorov2022resolution} compared multiple deep learning and Generative Adversarial Network (GAN)-based models under narrow versus wide masks—analogous to low- and high-octave corruptions—and reported a 3–5× deterioration in FID scores when transitioning from narrow to wide gaps. Similarly, Zhu et al.\ \cite{zhu2021image} evaluated a wide range of statistical, deep learning, and GAN-based inpainting methods under increasing mask rates (10–60\%) and showed that even the best-performing model experienced a 40\% reduction in PSNR (35→19) and a 20\% reduction in SSIM (0.99→0.87). Ma et al.\ \cite{ma2022regionwise} further demonstrated consistent degradation across deep learning and GAN-based approaches as mask rates increased from 10\% to 50\%, with SSIM decreasing from approximately 0.97 to 0.78 and PSNR from 33 to 23, and found that contiguous masks consistently resulted in worse performance than discontiguous ones.

\begin{figure}
    \centering
    \includegraphics[width=1\linewidth]{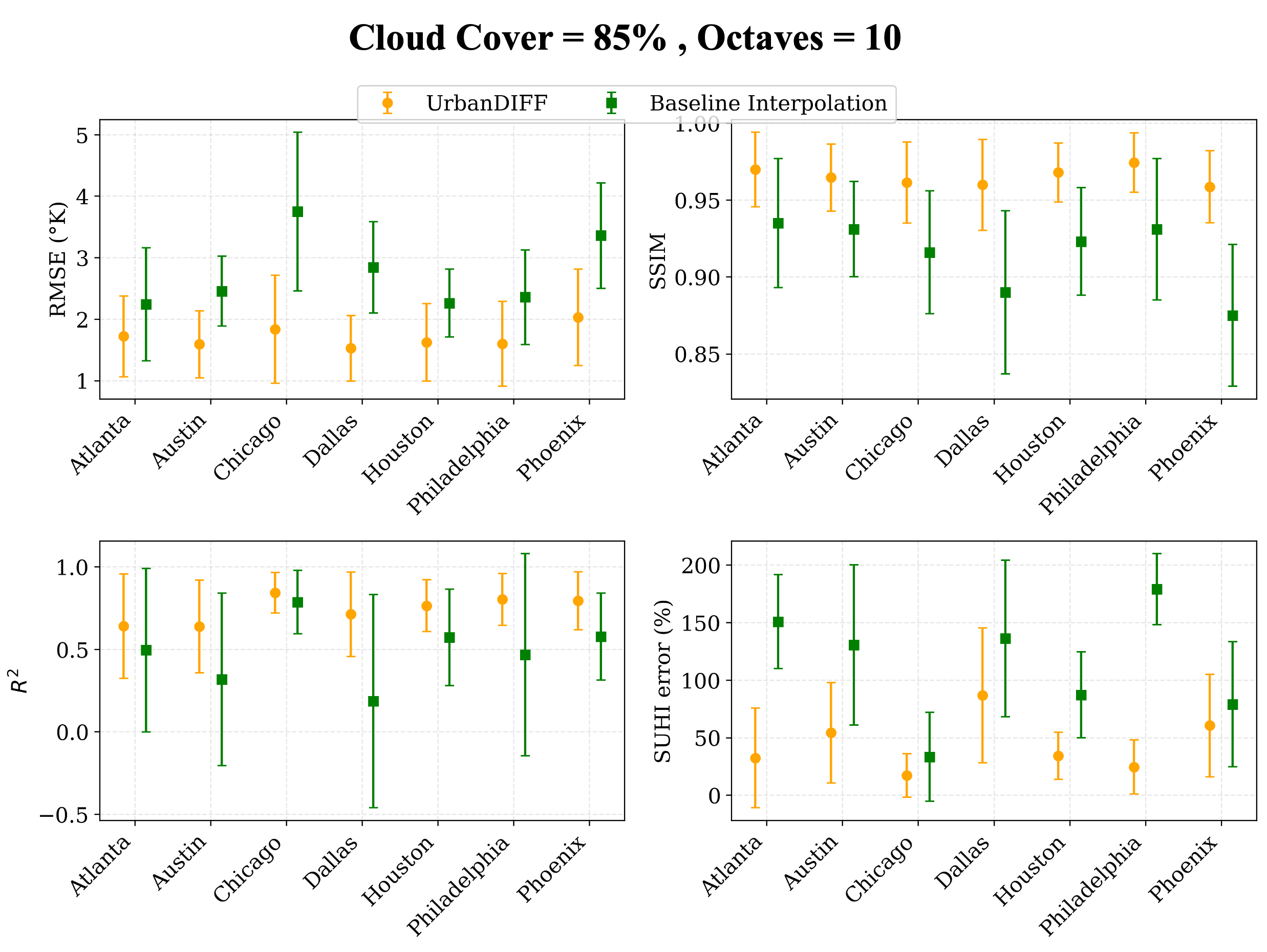}
   \caption{Comparison of UrbanDIFF reconstruction performance across studied urban regions under 85\% synthetic cloud coverage with octave 10, evaluated in terms of RMSE, SSIM, $R^2$, and SUHI error.}
    \label{fig:6}
\end{figure}

Figure~\ref{fig:6} complements Figures~4 and~5 by presenting a per-city comparison of UrbanDIFF and the interpolation baseline under severe cloud occlusion (85\% cloud cover, octave 10), thereby assessing the spatial generalization ability of UrbanDIFF across different urban regions. This analysis is particularly important because the availability of cloud-free or near–cloud-free training samples varies substantially across cities, which could otherwise bias reconstruction performance toward data-rich regions. All reported metrics are computed over held-out test images for each city, and error bars represent two standard deviations across test samples.
Across all evaluated cities, UrbanDIFF exhibits consistently strong and homogeneous performance. In terms of RMSE and SSIM, UrbanDIFF achieves mean RMSE values below 2~K and mean SSIM values above 0.95 for all cities, with relatively small standard deviations that are substantially lower than those of the interpolation baseline. 
UrbanDIFF also demonstrates consistent performance in terms of $R^2$, with mean values ranging approximately from 0.68 to 0.89 across cities and generally higher than those of the baseline. The variability of $R^2$ differs across regions, reflecting differences in the variance of the underlying LST fields rather than reconstruction instability. For example, Chicago exhibits lower $R^2$ variability, which is consistent with its higher intrinsic LST variance driven by strong land–water temperature contrasts associated with proximity to Lake Michigan. In contrast, cities such as Atlanta and Austin show higher $R^2$ variability, which is expected given their more spatially homogeneous LST fields and lower intrinsic variance.

In terms of SUHI estimation error, UrbanDIFF consistently outperforms the interpolation baseline across all cities, achieving both lower mean error and reduced variability. The mean SUHI error for UrbanDIFF ranges approximately from 35\% to 85\%, compared to 45\% to 140\% for the baseline. Lower SUHI error indicates improved preservation of urban–rural thermal gradients, which are critical for surface urban heat island analysis.
Among the evaluated cities, Chicago, Houston, and Philadelphia exhibit the lowest SUHI estimation errors, whereas Atlanta, Austin, and Phoenix show comparatively larger errors. This pattern is consistent with differences in urban–rural thermal contrast across cities. Cities such as Chicago, Philadelphia, and Houston are characterized by sharper urban–rural transitions, while cities such as Austin, Atlanta, and Dallas exhibit more gradual transitions between urban and surrounding rural areas. This behavior is supported by independent estimates of average daytime SUHI intensity for 2020 derived from MODIS Terra LST using the portal developed by Chakraborty and Lee \cite{chakraborty2019simplified}. Specifically, reported daytime SUHI intensities for Houston, Philadelphia, and Chicago are 3.24, 2.29, and 1.99~K, respectively, compared to lower values of 1.81, 1.76, and 1.40~K for Dallas, Austin, and Atlanta.
These results suggest that, irrespective of differences in training data availability across cities, UrbanDIFF is particularly effective at reconstructing urban LST fields in regions with pronounced urban–rural thermal contrasts, highlighting its ability to preserve urban thermal structure even under severe cloud occlusion. Overall, the consistent performance across heterogeneous urban regions demonstrates the robustness and generalizability of UrbanDIFF for purely spatial LST reconstruction under challenging cloud conditions.

\begin{figure}
    \centering
    \includegraphics[width=1\linewidth]{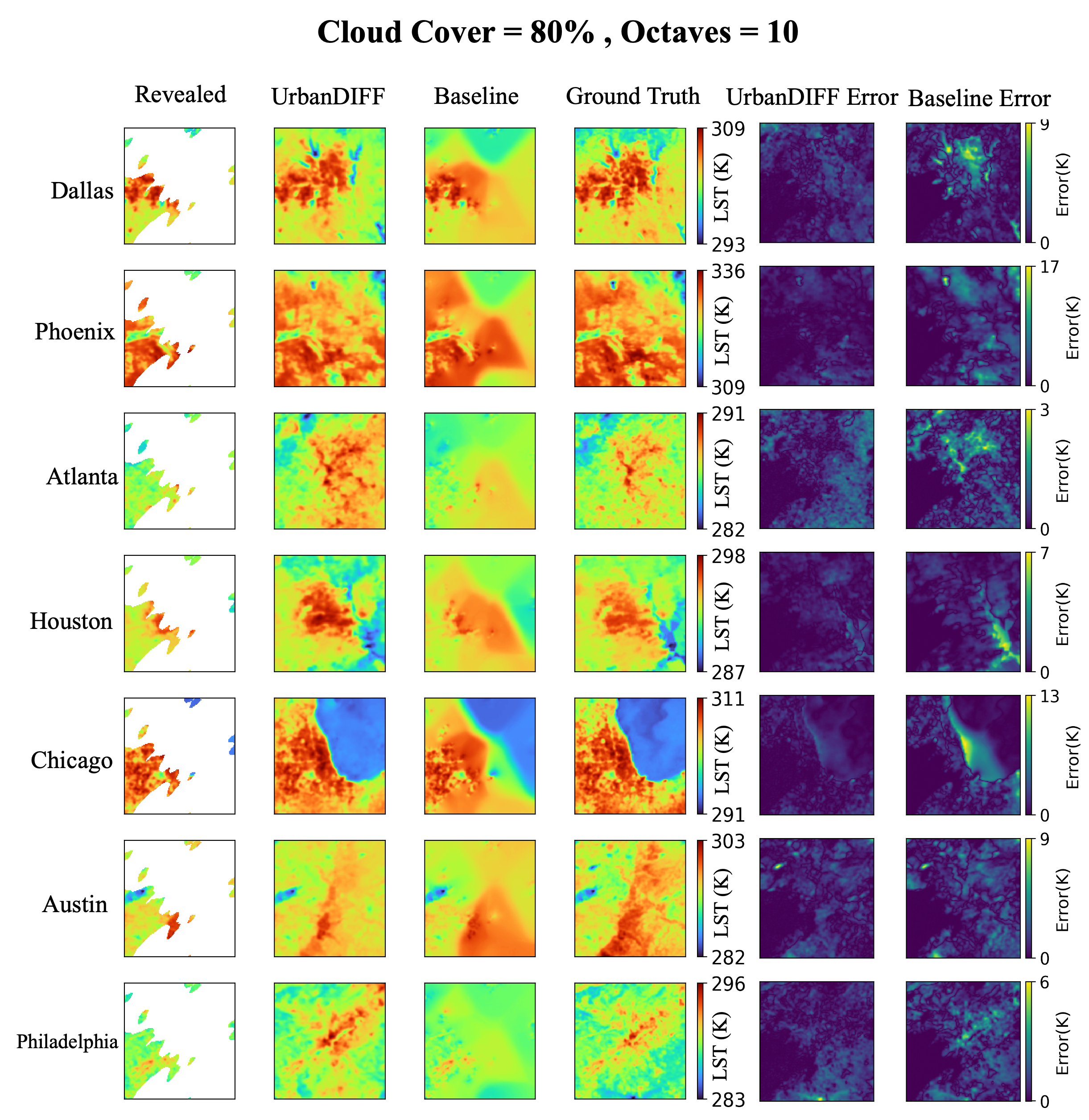}
    \caption{Visualization of LST reconstruction results from UrbanDIFF and the interpolation baseline under heavy synthetic cloud occlusion (80\% cloud coverage, octave 10), along with the corresponding error maps.}

    \label{fig:7}
\end{figure}

Finally, Figure~\ref{fig:7} provides a qualitative comparison of UrbanDIFF and the interpolation baseline for LST reconstruction under a heavy synthetic cloud occlusion (80\% cloud coverage, octave 10) across the studied urban regions. While the interpolation baseline produces overly smooth reconstructions that fail to recover fine-scale spatial structure—particularly across urban–rural boundaries—UrbanDIFF successfully reconstructs coherent spatial patterns that closely resemble the ground-truth LST fields. This difference is especially evident in cities with strong spatial heterogeneity, such as Chicago and Houston, where UrbanDIFF preserves sharp thermal gradients associated with land–water interfaces and dense urban cores. The corresponding error maps further highlight these differences: UrbanDIFF exhibits lower-magnitude and more spatially diffuse errors, whereas the baseline shows larger, spatially clustered errors aligned with missing regions. These qualitative results are consistent with the quantitative metrics reported earlier and illustrate UrbanDIFF’s ability to recover realistic urban thermal structure under severe cloud occlusion.

\section{Limitations}

UrbanDIFF is intentionally designed as a purely spatial, single-image gap-filling model. As a result, it relies on the presence of some revealed (cloud-free) pixels within each input scene. Under prolonged periods of persistent cloud cover, where little or no valid LST information is available, purely spatial reconstruction becomes ill-posed, and UrbanDIFF cannot recover physically meaningful temperature fields without additional information.

In its current formulation, UrbanDIFF does not incorporate temporal context or cross-sensor auxiliary variables (e.g., passive microwave observations or atmospheric reanalysis data). While this design choice enables isolation of the intrinsic capability of diffusion-based spatial modeling, it also limits performance in scenarios where temporal continuity or external observations are necessary to constrain reconstruction.

Like other deep learning–based generative models, UrbanDIFF also requires a sufficient number of cloud-free or near–cloud-free samples for training. In regions with persistently high cloud frequency or during nighttime conditions—where thermal contrast is reduced and missingness is more severe—the availability of suitable training data may be limited, potentially affecting generalization.

These limitations suggest natural directions for future work, including the integration of temporal information to extend UrbanDIFF toward more operational all-weather LST reconstruction.

\section{Conclusion}

In this study, we developed UrbanDIFF, a denoising diffusion–based framework for urban land surface temperature (LST) reconstruction under heavy cloud occlusion, with an explicit focus on preserving urban–rural thermal gradients critical for surface urban heat island (SUHI) analysis. Unlike most existing LST reconstruction approaches that rely on temporal information or multi-sensor data fusion, UrbanDIFF is formulated as a purely spatial gap-filling model.
UrbanDIFF incorporates two static conditioning variables—urban built-up surface data representing the city morphological skeleton and a digital elevation model (DEM)—to guide spatially coherent reconstruction. In addition, a pixel-guided refinement strategy is applied during inference, enforcing strict consistency between the generated samples and all revealed cloud-free pixels through supervised gradient-based updates. This design enables robust reconstruction under severe occlusion while maintaining fidelity to observed data.
The model was trained using MODIS Terra LST data spanning 2002–2025 over seven major U.S. metropolitan regions (Houston, Dallas, Austin, Atlanta, Phoenix, Chicago, and Philadelphia). Performance was evaluated on held-out test data using synthetic cloud masks with coverage ranging from 20\% to 85\% and varying spatial density. After hyperparameter optimization, UrbanDIFF was benchmarked against a widely used interpolation-based baseline.
Results demonstrate that UrbanDIFF consistently outperforms the baseline, with performance gains becoming more pronounced under higher cloud coverage and denser occlusion patterns. The model achieves robust reconstruction quality across all cities, largely independent of variations in training data availability. Notably, UrbanDIFF exhibits strong capability in estimating SUHI intensity, particularly in dense urban regions with sharp urban–rural transitions. Even under severe synthetic cloud occlusion conditions (85\% coverage, high-density masks), UrbanDIFF achieves average SSIM values above 0.95, RMSE below 2~K, $R^2$ exceeding 0.85, and SUHI error around 60\% across the evaluated regions.
Overall, these findings highlight the potential of denoising diffusion–based spatial modeling for reliable urban LST reconstruction under challenging cloud conditions. While UrbanDIFF is not intended as a complete operational replacement for spatiotemporal or multi-source methods, it provides a strong methodological foundation for future extensions that integrate temporal context, additional conditioning variables, or uncertainty-aware analysis for urban climate applications.

\bibliographystyle{unsrt}  
\bibliography{references}

\end{document}